\documentclass[runningheads]{llncs}

\usepackage[camera-ready,year=2024,ID=716]{eccv}

\usepackage{eccvabbrv}

\usepackage{times}
\usepackage{epsfig}
\usepackage{graphicx}
\usepackage{amsmath}
\usepackage{amssymb}
\usepackage{subcaption}
\usepackage{color}
\usepackage{booktabs}
\usepackage{mathastext}
\usepackage{multirow}
\usepackage{comment}
\usepackage{diagbox}
\usepackage{bbding}
\usepackage{appendix}
\usepackage{eso-pic}
\usepackage[normalem]{ulem}
\usepackage[accsupp]{axessibility}  

\usepackage[pagebackref,breaklinks,colorlinks,citecolor=eccvblue]{hyperref}

\usepackage{orcidlink}

\begin{document}

\title{NeRF2Points: Large-Scale Point Cloud Generation From Street Views' Radiance Field Optimization} 


\author{Peng Tu\inst{1} \and
Xun Zhou\inst{1} \and
Mingming Wang\inst{2} \and
Xiaojun Yang\inst{1} \and
Bo Peng\inst{1} \and
Ping Chen\inst{3} \and\\
Xiu Su$^{*}$\inst{5} \and
Yawen Huang$^{*}$\inst{4} \and
Yefeng Zheng\inst{4} \and
Chang Xu\inst{5}}

\authorrunning{Peng~Tu et al.}

\institute{Ruqi Mobility Inc., Guangzhou, China \and
GAC R\&D Center, Guangzhou, China \and
MicroBT Inc., Shenzhen, China \and
Jarvis Research Center, Tencent YouTu Lab, Shenzhen, China \and
University of Sydney, Australian}

\maketitle
\let\thefootnote\relax\footnotetext{${}^{*}$Corresponding author; Contact Email: yh.peng.tu@gmail.com}

\vspace{-5pt}
\begin{abstract}
Neural Radiance Fields (NeRF) have emerged as a paradigm-shifting methodology for the photorealistic rendering of objects and environments, enabling the synthesis of novel viewpoints with remarkable fidelity. 
This is accomplished through the strategic utilization of object-centric camera poses characterized by significant inter-frame overlap. 
This paper explores a compelling, alternative utility of NeRF: the derivation of point clouds from aggregated urban landscape imagery. 
The transmutation of street-view data into point clouds is fraught with complexities, attributable to a nexus of interdependent variables. 
First, high-quality point cloud generation hinges on precise camera poses, yet many datasets suffer from inaccuracies in pose metadata.  
Also, the standard approach of NeRF is ill-suited for the distinct characteristics of street-view data from autonomous vehicles in vast, open settings.  
Autonomous vehicle cameras often record with limited overlap, leading to blurring, artifacts, and compromised pavement representation in NeRF-based point clouds.  
In this paper, we present NeRF2Points, a tailored NeRF variant for urban point cloud synthesis, notable for its high-quality output from RGB inputs alone.  
Our paper is supported by a bespoke, high-resolution 20-kilometer urban street dataset, designed for point cloud generation and evaluation. 
NeRF2Points adeptly navigates the inherent challenges of NeRF-based point cloud synthesis through the implementation of the following strategic innovations: 
(1) Integration of Weighted Iterative Geometric Optimization (WIGO) and Structure from Motion (SfM) for enhanced camera pose accuracy, elevating street-view data precision.  
(2) Layered Perception and Integrated Modeling (LPiM) is designed for distinct radiance field modeling in urban environments, resulting in coherent point cloud representations.  
(3) Geometric-aware consistency regularization to rectify geometric distortions in sparse street-view data, confirming superiority of NeRF2Points through empirical validation.
\keywords{Neural radiance fields \and Point cloud generation \and Street views \and Self-driving}
\end{abstract}

\vspace{-5pt}
\section{Introduction}
\label{sec:intro}

\begin{figure*}[t]
\vspace{-5pt}
\centering
\setcounter{subfigure}{0}
\subfloat[NeRF's camera settings]{\includegraphics[width=5cm]{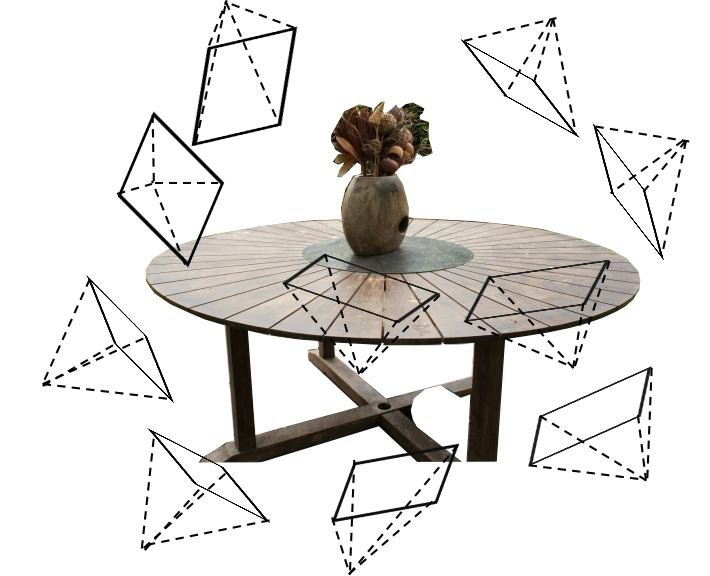}}\hspace{0.5mm}
\subfloat[Self-driving car's camera settings]{\includegraphics[width=6cm]{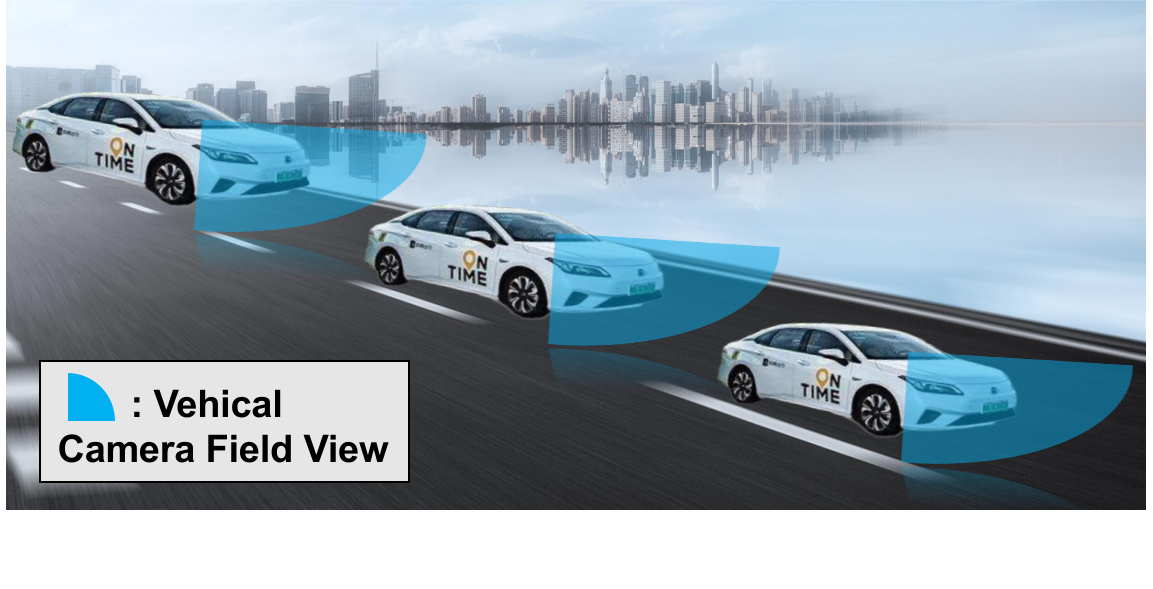}}\\\vspace{0.1mm}
\caption{
(a) NeRF's camera settings: the original NeRF is usually designed for object-centric scenes and requires hundreds of highly accurate camera poses to be heavily overlapped;
(b) Self-driving car's camera settings: Street view data are usually collected by using limited perspective cameras and run along the road, with almost no overlapping between different road sections and without object-centric camera views.
Moreover, some objects or contents only appear in limited images (2 $\sim$ 6) since the self-driving car moves fast.
However, most of the objects need to be reconstructed from hundreds of surrounding views with large overlaps, as shown in (a).
}
\label{camera setting}
\vspace{-5pt}
\end{figure*}

Neural Radiance Fields \cite{mildenhall2021nerf, deng2022depth, muller2022instant}(NeRF) have demonstrated exceptional capabilities in photorealistic rendering of novel viewpoints. As depicted in Fig. \ref{camera setting} (a), conventional NeRF architectures are tailored for object-centric scenarios, necessitating camera poses with extensive overlap to ensure accuracy. However, the data acquisition methodologies employed by autonomous vehicles yield street-view imagery that deviates from the stringent pose overlap requirements of traditional NeRF frameworks (Fig. \ref{camera setting} (b)), thereby constraining NeRF's applicability in the domain of autonomous navigation. Conversely, lidar-based point cloud acquisition remains a cornerstone for environmental perception within autonomous driving contexts. Yet, the prohibitive costs associated with lidar systems, coupled with the inherent sparsity of the resultant point clouds, present significant limitations. The prospect of generating point clouds from NeRF is particularly enticing, offering a cost-effective alternative that leverages readily available RGB data from vehicle-mounted cameras and holds the promise of producing denser point cloud reconstructions.

\begin{figure}[t]
\centering
\subfloat[Floating Artifacts~$\&$~Blurriness]{\includegraphics[width=5.5cm, height=2cm]{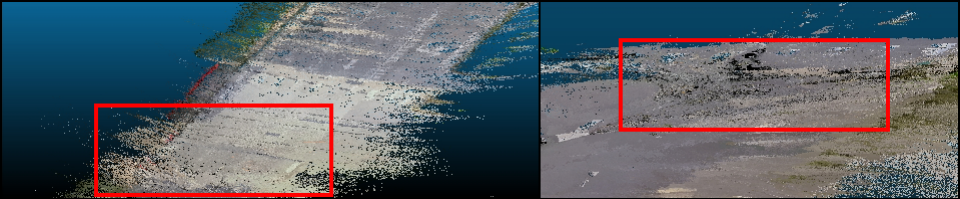}}\hspace{0.1mm}
\subfloat[Layering]{\includegraphics[width=5.5cm, height=2cm]{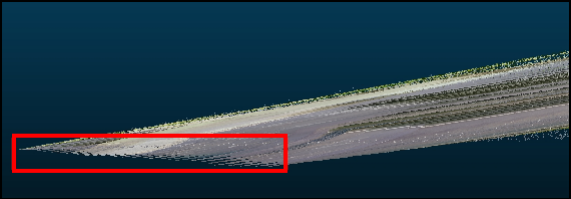}}\hspace{0.1mm}
\subfloat[Geometric Inconsistency]{\includegraphics[width=5.5cm, height=2cm]{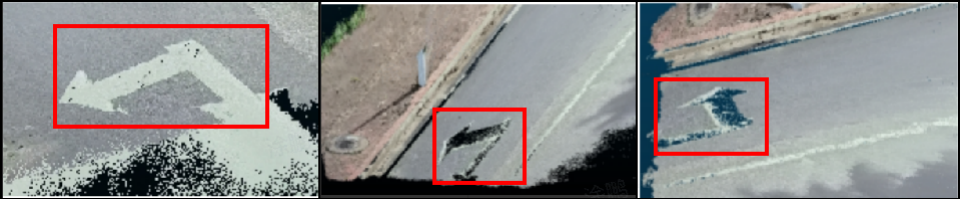}}\hspace{0.1mm}
\subfloat[Pavement Collapse]{\includegraphics[width=5.5cm, height=2cm]{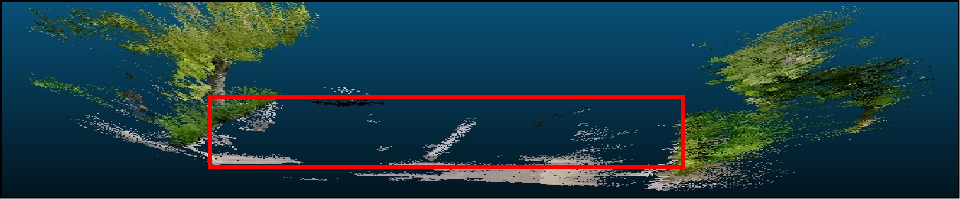}}\hspace{0.1mm}
\caption{
Four major defects while using NeRF to generate point clouds:
(a) and (c) is the floating artifacts $\&$ blurriness, and geometric inconsistency: These issues arise from the street-view data collection pipeline used in NeRF. 
The pipeline obtains sparse views, making it challenging to establish effective geometric constraints for modeling spatio-temporal radiance fields (as discussed in Sec. \ref{sec:GAC}).
(b) is the layering: Layering refers to the significant stratification observed between continuous but distinct road section point clouds generated by NeRF.
(d) is the pavement collapse: This phenomenon occurs due to the small and highly similar gradient values of weak texture pixels. 
As a result, recovering fine correspondent regions in the radiation field becomes difficult. 
Specifically, bundles of points representing road surfaces suffer from poor optimization, leading to inaccurate depth estimation in these regions.}
\label{pc_error}
\end{figure}

In summary, leveraging NeRF for the generation of point clouds from street-view imagery, as opposed to traditional lidar systems, is not only a more cost-effective and pragmatic solution but also facilitates the acquisition of high-quality, dense point clouds. These point clouds are invaluable for enhancing driving recognition algorithms, simulation fidelity, and the efficiency of data annotation processes.  
However, the task of generating point clouds via NeRF is fraught with formidable challenges:  
(1) The production of quality point clouds is predicated on the availability of highly accurate camera poses. In the context of novel view synthesis, the integrity of each pixel in the rendered image is contingent upon the precise weighted integration of points sampled along a ray traversing the neural radiance field. 
Consequently, while noisy poses may not compromise the structural integrity of novel view images, rendering many noisy datasets viable for NeRF-based exploration \cite{shi2022temporal, kim2023learning}, they do precipitate the generation of structurally deficient point clouds, as evidenced by the layering and blurring issues illustrated in Fig. \ref{pc_error} (a) and (b). 
This is attributable to the discontinuities introduced in the spatial uniformity of points and the suboptimal optimization process. 
Moreover, ROME \cite{mei2023rome} and S-NeRF \cite{xie2023s} suggest that datasets like KITTI \cite{Geiger2013IJRR} and nuScenes \cite{caesar2020nuscenes} benefit from pose re-optimization to yield superior 3D reconstruction outcomes. Thus, the primary challenge in NeRF-derived point cloud generation lies in the scarcity of street-view datasets with sufficiently precise camera poses.  
(2) Data harvested by autonomous vehicles encompass unbounded outdoor environments, where the camera's orientation and the scene's content may vary unpredictably, akin to the conditions present in the nuScenes and KITTI datasets. This introduces additional complexity to the data collection process, as the camera may capture imagery in any direction and at varying distances.
Data from limited-perspective cameras, which traverse roads with minimal view overlap and lack object-centric perspectives (Fig. \ref{camera setting} (b)), often result in geometric inconsistencies, floating artifacts, and pavement collapse in NeRF-generated point clouds (Fig. \ref{pc_error}).  
Objects often appear in few images (about 2 $\sim$ 6 views) due to the speed of vehicle, contrasting with the need for reconstruction from numerous overlapping views \cite{dai2017scannet,barron2022mip}.  MipNeRF-360 \cite{barron2022mip} addresses unbounded scene challenges by employing non-linear scene parameterization, online distillation, and distortion-based regularization, showing promise in extensive, borderless outdoor scenes.  
Block-NeRF \cite{tancik2022block} attempts to mitigate sparse and non-overlapping street data issues by using a multi-camera system to ensure sufficient overlap, yet this approach is neither cost-effective nor does it address the scarcity of object-centric views.  Despite these efforts, NeRF applications to street-view datasets still frequently result in blurriness and artifacts (Fig. \ref{pc_error}).

In this paper, we introduce NeRF2Points, a novel framework devised to surmount four prevalent challenges (Fig. \ref{pc_error}) in NeRF-generated point clouds:  
(1) Our dataset encompasses over 20 kilometers of meticulously captured street-view data from autonomous vehicles, complete with high-fidelity images, precise camera poses, depth maps, normal, and point clouds, which are instrumental in averting layering artifacts across disparate road sections.  
(2) We present the Layered Perception and Integrated Modeling (LPiM) strategy to address the issue of pavement collapse due to road texture deficiencies, effectively segmenting and reconciling road and street scene point clouds through calculated projection matrices from intersecting rays.  
(3) To combat the sparsity-induced artifacts, such as floaters, blurriness, and geometric inconsistencies, we have engineered two novel loss functions that promote spatial dynamism and temporal invariance consistency.

\vspace{-5pt}
\section{Related Works}
\subsection{3D Reconstruction}
3D reconstruction is the process of deducing the three-dimensional structure of an object or scene using multiple images captured from varied camera positions. 
Traditional reconstruction and novel view rendering \cite{agarwal2011building} often rely on structure-from-motion \cite{schonberger2016structure, losasso2004geometry} (SfM) and multi-view stereo \cite{seitz2006comparison, furukawa2015multi} (MVS) techniques \cite{agarwal2011building}. 
SfM is a technique used to reconstruct 3D models of scenes from a collection of 2D images taken from different angles. 
It involves estimating the camera poses and 3D structure of the scene simultaneously, using a bundle adjustment algorithm \cite{schonberger2016structure}. 
SfM is a fundamental technique that forms one of the cornerstones of vision-based 3D reconstruction, localization, and navigation systems. 
MVS algorithms rely heavily on the extraction and matching of feature points. 
However, the performance of these algorithms is limited in textureless scenes, such as road surfaces, where feature points are sparse and unevenly distributed \cite{seitz2006comparison, schonberger2016pixelwise}.

Deep learning based approaches have been widely used in the 3D scene and object reconstruction \cite{sitzmann2019deepvoxels, engelmann2021points}. 
They encode the feature through a deep neural network and learn various geometry representations, such as voxels \cite{kar2017learning, sitzmann2019deepvoxels}, patches \cite{groueix2018atlasnet} and meshes \cite{wang2018pixel2mesh}.
For example, DeepVoxels \cite{sitzmann2019deepvoxels} is a 3D-structured neural scene representation that encodes the view-dependent appearance of a 3D scene without having to explicitly model its geometry. 
It is based on a Cartesian 3D grid of persistent features that learn to make use of the underlying 3D scene structure. 
DeepVoxels is supervised, without requiring a 3D reconstruction of the scene, using a 2D re-rendering loss and enforces perspective and multi-view geometry in a principled manner. 
It combines insights from 3D computer vision with recent advances in learning image-to-image mappings. 
DeepVoxels is useful for generating a series of coherent views of the same scene, which is challenging for existing models that are based on a series of 2D convolution kernels.

\vspace{-5pt}
\subsection{Neural Radiance Fields}
Neural Radiance Fields (NeRF) is proposed in \cite{mildenhall2021nerf} as an implicit neural representation for novel view synthesis.
Various types of NeRFs haven been proposed for acceleration \cite{yu2021plenoctrees, rebain2021derf}, better generalization abilities \cite{yu2021pixelnerf, trevithick2021grf}, new neural signed distance function \cite{yariv2021volume, wang2021neus}, large-scale scenes \cite{tancik2022block, zhang2020nerf++}, and depth supervised training \cite{deng2022depth, rematas2022urban}.

\vspace{-5pt}
\subsubsection{Large-scale Scenes} ~~Many NeRF have been proposed to address the challenges of large-scale outdoor scenes \cite{martin2021nerf, li2021neural, rematas2022urban}.
A simple example can be drawn from NeRF in the Wild \cite{martin2021nerf}.
This method is a deep learning-based method for synthesizing novel views of complex outdoor scenes using only unstructured collections of in-the-wild photographs. 
It overcomes the limitations of NeRF by modeling the appearance variations of outdoor scenes by learning low-dimensional space that captures the appearance variations of the scene and using it to synthesize novel views of the scene. 

\vspace{-5pt}
\subsubsection{Signed Distance Function} 
NeuS \cite{wang2021neus} is a representative method of the signed distance function (SDF).
It is used for the neural surface reconstruction that reconstructs objects and scenes with high fidelity from 2D image inputs by using a new volume rendering method to train a neural signed distance function representation of the surface, which is represented as the zero-level set of the SDF. 
This method overcomes the limitations of existing neural surface reconstruction approaches, such as DVR \cite{niemeyer2020differentiable} and IDR \cite{yariv2020multiview}, which require foreground masks as supervision and struggle with the reconstruction of objects with severe self-occlusion or thin structures. 

\vspace{-5pt}
\subsubsection{Depth Supervision}
Parts of NeRF \cite{deng2022depth, xu2022point, neff2021donerf} use deep supervision to overcome the failure of their depth estimation.
DS-NeRF \cite{deng2022depth} utilizes the sparse depth generated by COLMAP \cite{schonberger2016structure} to supervise the NeRF training. 
PointNeRF \cite{xu2022point} uses point clouds to boost the training and rendering with geometry constraints. 
DoNeRF \cite{neff2021donerf} realizes the ray sampling in a log scale and uses the depth priors to improve
ray sampling.

However, these methods are designed for the novel view synthesis processing of small-scale objects \cite{wang2021neus} or indoor scenes \cite{barron2022mip}. 
Generating point clouds from NeRF remains unexplored.

\vspace{-5pt}
\section{NeRF To Point Clouds}
\subsection{Preliminary}
Neural Radiance Fields (NeRF) is a continuous 5D scene representation with RGB colors and volume density. 
It was learned by a multi-layer perceptron (MLP) that links every image pixel to special 3D coordinates $\mathbf{x} = (x, y, z)$ with its viewing direction $\hat{\mathbf{d}}=(\theta, \phi)$ which is a normalized 3D Cartesian unit vector:
\begin{equation}
\digamma(\gamma(\mathbf{x}), \gamma(\hat{\mathbf{d}})) \to (\mathbf{c}(r, g, b), \mathbf{\sigma}),
\label{nerf_mapping}
\end{equation}
where the MLP $\digamma(.)$ maps the input 5D coordinates to corresponding volume density $\mathbf{\sigma}$ and directional emitted color $\mathbf{c}$. 
Moreover, $\gamma(.)$ represents the positional encoding function to insert positional information to the raw coordinates of the input points in order to induce the network to learn higher-frequency features \cite{mildenhall2021nerf, lin2021barf}.

NeRF compositing RGBA color along an input ray $\mathbf{r}(t)=\mathbf{o}+t\mathbf{d}$, where $\mathbf{o}$ denotes the camera origin and $\mathbf{d}=\left\| \mathbf{d} \right\|_2 * \hat{\mathbf{d}}$ is unnormalized direction vector \cite{max1995optical}.
RGBA stands for red, green, blue, and alpha which indicates how opaque each pixel is and allows an image to be combined over others using alpha compositing, with transparent areas and anti-aliasing of the edges of opaque regions \cite{urban2019redefining}.
Rendered RGB color $\hat{\mathbf{c}}$ are obtain by:
\begin{equation}
\hat{\mathbf{c}}(\mathbf{r}) = \int_{t_{n}}^{t_{f}} T(t) \sigma(\mathbf{r}(t)) \mathbf{c}(\mathbf{r}(t), \hat{\mathbf{d}}) ~dt,
\label{nerf_rgb_redering}
\end{equation}
where $t_n$, $t_f$ indicates the nearest and farthest sampling plane along rays.
$T(t) = exp(- \int_{t_n}^{t} \sigma(s) ~ds)$ is the degree of transparency.
Eq. \ref{nerf_rgb_redering} approximated RGB values with numerical quadrature by sampling points along rays.
Finally, NeRF can be trained with minimize the mean squared error (MSE) function between rendered RGB color $\hat{\mathbf{c}(\mathbf{r})}$ and ground-truth pixels of $\mathbf{c}(\mathbf{r})$ by follows:
\begin{equation}
\ell_{MSE} = \sum_{\mathbf{r}\in\mathcal{R}} \left\| \hat{\mathbf{c}}(\mathbf{r}) - \mathbf{c}(\mathbf{r}) \right\|_2^2,
\label{nerf_rgb_loss}
\end{equation}
where $\mathcal{R}$ are bundles of rays uniform sampling from a set of images.

\begin{figure}[htp]
\centering
\includegraphics[width=12cm,]{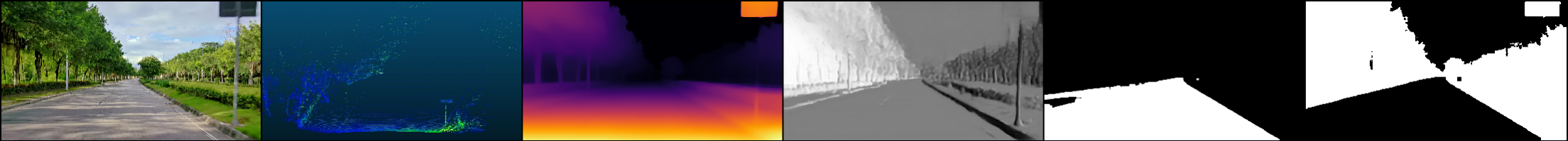} 
\includegraphics[width=12cm,]{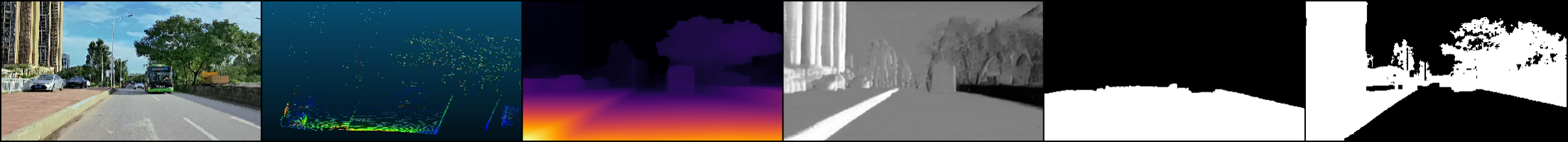} 
\includegraphics[width=12cm,]{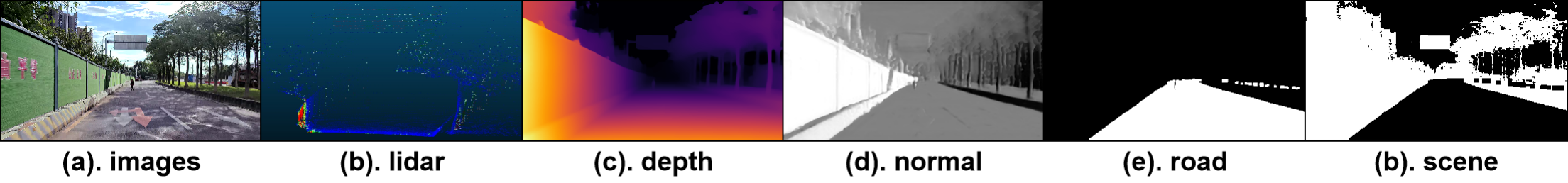} 
\caption{Here are some examples of the 20 kilometers of street view data we’ve collected.}
\label{exm_collectdata}
\end{figure}

\vspace{-5pt}
\subsection{Build Up NeRF2Points}
\subsubsection{Street View Data Collection}\label{sec:svdc}
We collect over 20 kilometres of about 180k high-definition images and correspondent point clouds (obtained from lidar sensor) of street view data from self-driving cars through six cameras, including front $\&$ front right $\&$ front left $\&$ back $\&$ back right $\&$ back left cameras.
These cameras have 3840 $\times$ 2160 resolution except the back camera, whose resolution is half (1920 $\times$ 1080) that of others.
Further, we obtain camera initial poses by using the Wheel-IMU-GNSS odometry (WIGO) algorithm \cite{lee2020visual, gtsam}, which combines sensor readings with pose graph optimization \cite{gtsam, carlone2015initialization} (PGO).
When the WIGO algorithm provides a descent global 6-DoF \cite{sundermeyer2021contact} (Degree of Freedom) pose with a real-world scale by combining GNSS signals with pose graph optimization, the IMU and wheel sensor provide relative pose constraints between consecutive frames. 
Related scale information mainly comes from the GNSS observations. 
We treated WIGO results as initial camera poses and used them as inputs of the Structure from Motion \cite{schonberger2016structure, losasso2004geometry} (SfM) to refine these camera poses.
In detail, we use COLMAP \cite{schonberger2016structure} with SfM to estimate highly accurate camera parameters and a sparse point cloud from given street view data with coarse camera poses.
Additionally, we derived depth and normal estimations from omnidata \cite{eftekhar2021omnidata} and extracted road and other regions in images using Mask2Former \cite{cheng2022masked}. 
Each single-frame image captured by a camera provides five corresponding pieces of information: ridar point clouds, depth and normal estimations, and masks for road surfaces and other regions, as shown in Fig. \ref{exm_collectdata}. 
While the ridar collected point cloud is included in our dataset, it will not be used for radiation field training. 
Its primary purpose is to serve as ground truth for evaluating the accuracy of the point clouds generated by the radiation field.

\vspace{-5pt}
\begin{figure*}[htp]
\centering
\includegraphics[width=11cm,]{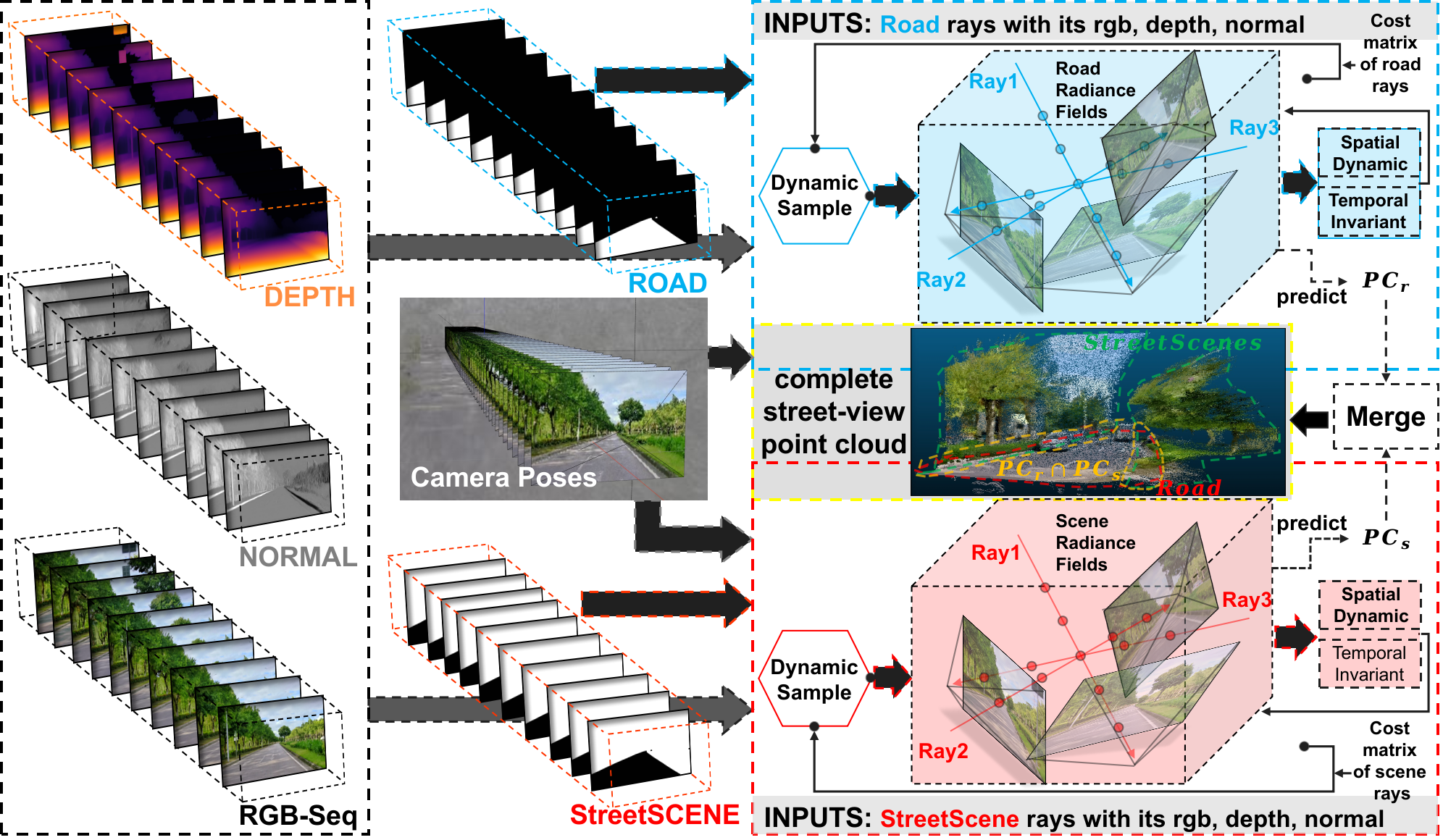} 
\caption{Overview of our proposed method: generation point clouds from NeRF (NeRF2Points). 
RGB sequences and their corresponding depth maps, along with normal vectors, are fed into the ray gate function $\textit{M}(.)$. 
This process separates the information related to road and street scenes for each ray.
After individually modeling the point clouds for road and street scenes, we merge them to create a complete street-view point cloud. 
The resulting merged point cloud is highlighted by the yellow dotted box in the center of the right image.}
\label{NeRF2Points}
\end{figure*}

\vspace{-5pt}
\subsubsection{Layered Perception and Integrated Modeling (LPiM)}
Handling sparse views in NeRF poses a genuine challenge when aiming to generate complete and precise point clouds.
Illustrated in Fig. \ref{pc_error} (d), our attempts to generate point cloud data for scene representation using NeRF with pure RGB images reveal that radiance fields exhibit a preference for high-frequency features (such as buildings or trees), yet struggle to accurately recover subtle texture details like road surfaces. 
Weak texture recovery has always been a long-standing challenge in 3D reconstruction. 
The image gradient plays a crucial role in all 3D reconstruction tasks, as it provides critical information to distinguish whether two points represent the same location \cite{yang2021practical}. 
However, when dealing with weak textures, their gradient values are very small and highly similar, making it difficult to recover fine texture details in 3D reconstruction \cite{fu2022weak, tang2023delicate}.
This limitation also contributes to the phenomenon of pavement collapse in point clouds generated by NeRF. 
Specifically, bundles of points representing road surfaces suffer from poor optimization, resulting in terrible depth estimation for these regions.
To overcome this challenge, we design layered perception and integrated modeling (LPiM) for radiance field modeling. 
Refer to Fig. \ref{NeRF2Points} for an illustration.
Layered perception adopts a divide-and-conquer approach, separately modeling the road surface and street view radiation fields. 
This strategy effectively addresses the issue of pavement collapse during point cloud generation by ensuring that gradient optimization of weak textures is not overshadowed by high-frequency details. 
Additionally, this method enables MLP within radiance fields to facilitate extensive interaction between co-frequency domain information, which can enhance the feature robustness of weak textures.  
Layered perception yields complete but separate point clouds, including road surfaces and street scenes.
Integrated modeling enables spatially separated point clouds to be merged into a complete street-view point cloud by calculating the projection matrix from shared rays.

In detail, assuming a bundle of rays from an image $\mathbf{I} \in \mathbb{R}^{H \times W \times 3}$, each ray $\mathbf{Ray}_{i}=\{\mathbf{x}_{i}, \mathbf{d}_{i}\}$ where $\mathbf{x}_{i}=(x_{i}, y_{i}, z_{i}), i \in \{0, 1, 2, ..., H \times W\}$ and $\mathbf{d}_{i}=(\theta_{i}, \phi_{i})$ are its origin and viewing direction, respectively.
We feed image $I$ to a ray gate $\mathbf{\textit{M}}(.)$ (segmentation model \cite{cheng2022masked} trained on Cityscapes \cite{Cordts2016Cityscapes}), then obtain two sets of reflection $\mathbf{Ray}_{i_{1}}^{r}$, $\mathbf{Ray}_{i_{2}}^{s}$ from road and scene surface:
\begin{equation}
\begin{aligned}
\mathbf{Ray}_{i_{1}}^{r} &= \mathbf{Ray}_{i}(\mathbf{\textit{M}}(\mathbf{I})),\\
\mathbf{Ray}_{i_{2}}^{s} &= \mathbf{Ray}_{i}(1-\mathbf{\textit{M}}(I)),\\
\mathbf{Ray}_{i_{3}} &= \mathbf{Ray}_{i_{1}}^{r} \cap \mathbf{Ray}_{i_{2}}^{s};\\
\textit{s.t.}~~i_{1}, i_{2}, i_{3} &\in \{0, 1, 2, ..., H \times W\},~~\&\\
\mathbf{Ray}_{i}=(\mathbf{Ray}_{i_{1}}^{r} - &\mathbf{Ray}_{i_{3}}) \cup (\mathbf{Ray}_{i_{2}}^{s} - \mathbf{Ray}_{i_{3}}) \cup \mathbf{Ray}_{i_{3}}.
\label{rays_of_road_scene}
\end{aligned}
\end{equation}
Within the function of the ray gate $\mathbf{\textit{M}(.)}$, the embedded morphological dilation operator facilitates the intersection of the road surface and street scene along their edges.
The intersection of ray bundles $\mathbf{Ray}_{i_{1}}^{r}$ and $\mathbf{Ray}_{i_{2}}^{s}$ is $\mathbf{Ray}_{i_{3}}$. 
We intend to leverage this prior knowledge of shared $\mathbf{Ray}_{i_{3}}$ for seamlessly stitching the point clouds representing road surfaces and street scenes.
In various scenes, we captured the ground truth information for reflected rays, including color ($\mathbf{c}_{i_{1}}^{r}, \mathbf{c}_{i_{2}}^{s}$), depth ($\mathbf{t}_{i_{1}}^r, \mathbf{t}_{i_2}^s$), and normal vectors ($\mathbf{n}_{i_1}^r, \mathbf{n}_{i_2}^s$).
These prior observations will act as the supervisory signal, guaranteeing the successful training of radiance fields, as detailed in Sec. \ref{sec:total_loss}.
Following the output of radiance fields, which generate separate point clouds for the road surface and street scene, we merge these point clouds to create a cohesive street-level representation.
We represent the output point clouds for the road surface and street scene as $\mathbf{PC}_r$ (with dimensions $N_1 \times 3$) and $\mathbf{PC}_s$ (with dimensions $N_2 \times 3$), where $N_1$ and $N_2$ denote the respective point counts.
Given the intersection points of two point clouds $\mathbf{P} = \{\mathbf{p}_1, \mathbf{p}_2, ..., \mathbf{p}_{N^{'}}\} \subset \mathbf{PC}_r, \mathbf{Q} = \{\mathbf{q}_1, \mathbf{q}_2, ..., \mathbf{q}_{N^{'}}\} \subset \mathbf{PC}_s ~~(N^{'} << N_{1}, N_{2})$ respectively, we employ  a rigid transformation \cite{besl1992method} on $\mathbf{P}$ to align to $\mathbf{Q}$ by using a rotation matrix $\mathbf{R} \in \mathbb{R}^{3 \times 3}$ and a translation vector $\mathbf{t} \in \mathbb{R}^{3}$ :
\begin{equation}
\min_{\mathbf{R}, \mathbf{t}} ~~ \sum_{i=1}^{N^{'}} \bigl(\textit{Dis}_i(\mathbf{R}, \mathbf{t}) \bigr)^2 + \textit{I}_{SO} (\mathbf{R}),
\label{rigid_transformation}
\end{equation}
where $\textit{Dis}_i(\mathbf{R}, \mathbf{t}) = \min_{\mathbf{q}_{i} \in Q, \mathbf{p}_{i} \in P} \| \mathbf{R}\mathbf{p}_{i} + \mathbf{t} - \mathbf{q}_{i}\|$ is the distance from the transformed point $\mathbf{R}\mathbf{p}_{i} + \mathbf{t}$ to the target point $\mathbf{q}_{i}$, and the $\textit{I}_{SO}(.)$ is an indicator function for the special orthogonal group $SO$, which requires $\mathbf{R}$ to be a rotation matrix:
\begin{equation}
\textit{I}_{SO}(\mathbf{R})
	=
	\left\{
	\begin{array}{ll}
	0, & \textrm{if}~ \mathbf{R}^T \mathbf{R} = \mathbf{I} ~\textrm{and}~\det(\mathbf{R}) = 1,\\
	+\infty, & \textrm{otherwise}.
	\end{array}
	\right.
\end{equation}
After obtaining the converged rotation $\mathbf{R}$ and translation $\mathbf{t}$ matrix, we produced the cohesive street-level point clouds $\mathbf{PC} \in \mathbb{R}^{(N_1 + N_2) \times 3}$ by:
\begin{equation}
\mathbf{PC} = \mathbf{PC}_r \mathbf{R} + \mathbf{t} + \mathbf{PC}_s.
\end{equation}

\vspace{-5pt}
\subsubsection{Geometric-Aware Consistency Regularization (GAC)}\label{sec:GAC}
LPiM focuses on preventing pavement collapse in point clouds of road surfaces. 
However, it does not adequately tackle the challenges arising from inconsistent geometry and floaters during point cloud generation.
These issues stem from the street-view data collection pipeline, which obtains sparse views, making it difficult to establish effective geometric constraints for modeling spatio-temporal radiance fields.
On-boat cameras, while capturing data at a fixed angle as they move along the road, introduce inductive biases that render the radiance field sensitive to viewing angle shifts and create cumulative depth estimation errors.
The radiance fields’ sensitivity to viewing angle changes results in geometric inconsistencies of generated point clouds.
As an illustration, road markings are clearly visible from the initial viewing angle of the training data.
However, if the viewing angle shifts, the previously identifiable area information within the point cloud becomes indistinct (Fig. \ref{pc_error} (c)). 
Additionally, the cumulative depth estimation errors result in irregular floaters accumulating above the road surface (see Fig. \ref{pc_error} (a)).
To address these challenges, we propose adding geometric consistency constraints during the spatial-temporal modeling process of radiance fields. Specifically, we introduce spatial dynamic resample consistency and temporal invariant feature consistency.

\begin{figure}[htp]
\centering
{\includegraphics[width=6.5cm]{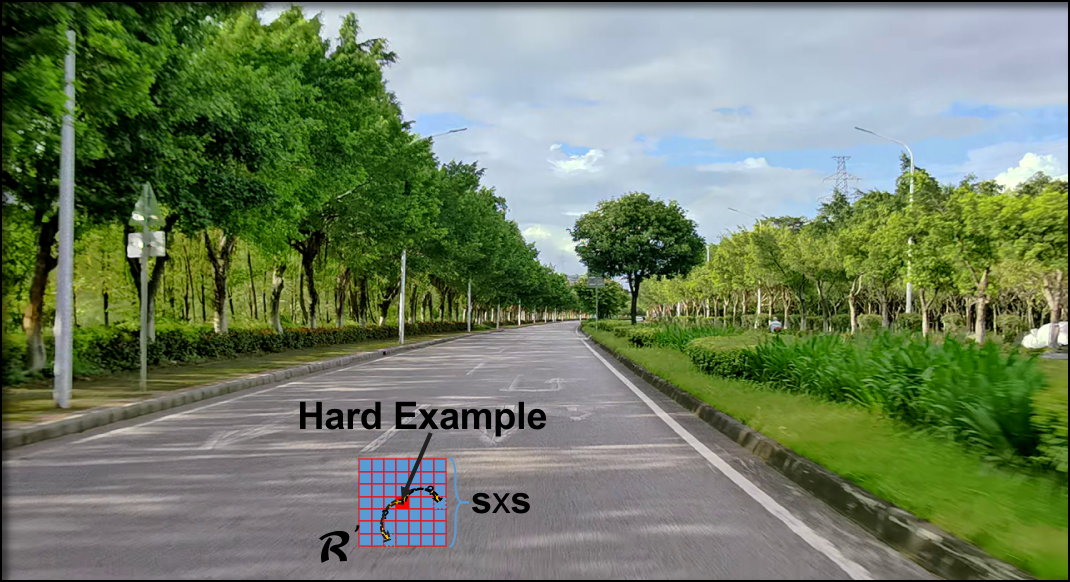}}\hspace{0.1mm}
\caption{
We focus on optimizing hard examples to mitigate artifacts in generated point clouds.
In the above picture, the red and blue grid corresponds to the $s ~\times~ s$ field $\mathcal{R}^{'}$, centered around the red hard sample.
Within this field, we randomly choose sample points and red hard sample points to create a consistency loss, which helps control the error optimization of the hard sample points.}
\label{sdc}
\end{figure}
\textbf{Spatial Dynamic Consistency:}
Within the framework of learning-based methods, erroneous optimization of hard samples can result in substantial matching overhead and give rise to the aforementioned artifacts in the generated point clouds. 
To tackle this challenge, our approach involves identifying the ray with the highest matching cost within each ray bundle and treating it as a complex optimization sample. 
Furthermore, we impose consistent representation constraints in the spatial neighbourhood of each challenging sample, as shown in Fig. \ref{sdc}.
Record the ray bundles coming from the road  surfaces $\mathbf{Ray}_{i_{1}}^{r}$ or street scenes $\mathbf{Ray}_{i_{2}}^{s}$ as $\mathbf{Ray}_{i}^{'}$.
Their matching cost $\mathbf{cos}_{i} \in \mathbb{R}^{k}$ ($k=len(\mathbf{Ray}_{i}^{'})$) used to determine which ray is a complex optimization sample can be defined as the weighted summation of their optimization error, including the losses of depth $\ell_{depth}(.)$, RGB $\ell_{rgb}(.)$, and normal $\ell_{normal}(.)$:
\begin{equation}
\begin{aligned}
\mathbf{cos}_{i} &= \lambda_{1} \ell_{depth}(.) + \lambda_{2} \ell_{rgb}(.) + \lambda_{3} \ell_{normal}(.),\\
\ell_{depth}(\mathbf{t}_{1}^{'}, \mathbf{t}_{2}^{'}) &= \ell_{MSE}(\textit{norm}(\mathbf{t}_{1}^{'}), \textit{norm}(\mathbf{t}_{2}^{'}))=\sum_{} \left\| \textit{norm}(\mathbf{t}_{1}^{'}) - \textit{norm}(\mathbf{t}_{2}^{'}) \right\|_2^2,\\
\ell_{rgb}(\mathbf{c}_{1}^{'}, \mathbf{c}_{2}^{'}) &= \ell_{MSE}(\mathbf{c}_{1}^{'}, \mathbf{c}_{2}^{'})=\sum_{} \left\| \mathbf{c}_{1}^{'} - \mathbf{c}_{2}^{'} \right\|_{2}^{2},\\
\ell_{normal}(\mathbf{n}_{1}^{'}, \mathbf{n}_{2}^{'}) &= \sum_{} {\lVert \mathbf{n}_{1}^{'} - \mathbf{n}_{2}^{'} \rVert}_1 +  {\lVert  1 - \mathbf{n}_{1}^{'\top}  \mathbf{n}_{2}^{'} \rVert}_1,
\label{cost}
\end{aligned}
\end{equation}
where $\lambda_{1}, \lambda_{2}, \lambda_{3}$ stands for the balanced coefficient fixed by experimentation, and ($\mathbf{c}_{1}^{'}, \mathbf{c}_{2}^{'}$), ($\mathbf{t}_{1}^{'}, \mathbf{t}_{2}^{'}$), ($\mathbf{n}_{1}^{'}, \mathbf{n}_{2}^{'}$) are the RGB values, depth maps, normal vectors of $\mathbf{Ray}_{i,n}^{'}$ correspondent ground-truth and predictions.
Besides, $\textit{norm(.)}$ is max-min normalization function.
Next, we rank the cost $\mathbf{cos}_{i}$ and select the top-$\textit{n}$ highest value corresponding rays $\mathbf{Ray}_{i,n}^{'}$: 
\begin{equation}
\mathbf{Ray}_{i,n}^{'} = \mathbf{Ray}_{i}^{'}[\textit{Rank}(c_{i}).top(\textit{n})].
\end{equation}
Regarding the image pixel indices of each ray from $\mathbf{Ray}_{i,n}^{'}$ are $\mathbf{ind}$, then we construct proposal regions $\mathcal{R}^{'}$ whose size is $s \times s$ with each index from $\textit{ind}$ as the central point.
The consistent matching samples of each ray in $\mathbf{Ray}_{i,n}^{'}$ will be randomly selected in $\mathcal{R}^{'}$, while the selected samples in $\mathcal{R}^{'}$ will not be fixed among different optimization step.
We calculate the spatial dynamic consistency loss $\ell_{sdc}$ by:
\begin{equation}
\begin{aligned}
\ell_{sdc} &= \lambda_{1}^{'} \ell_{jsd} (\phi(\mathbf{r}_{1}^{'}),\phi(\mathbf{r}_{2}^{'})) + \lambda_{2}^{'} \ell_{rgb}(.) + \lambda_{3}^{'} \ell_{depth}(.),\\
\ell_{jsd} (\phi(\mathbf{r}_{1}^{'}),\phi(\mathbf{r}_{2}^{'}))  &= 0.5 * \frac{\sum \phi(\mathbf{r}_{1}^{'}) * ln(\phi(\mathbf{r}_{2}^{'})) + \sum \phi(\mathbf{r}_{2}^{'}) * ln(\phi(\mathbf{r}_{1}^{'}))}{(\phi(\mathbf{r}_{1}^{'}) + \phi(\mathbf{r}_{2}^{'}))/2},\\
\phi(\mathbf{x}) &= \frac{e^{x_{i}}}{\sum_{c=1}^{C} e^{x_c}},
\label{sdc}
\end{aligned}
\end{equation}
where $\mathbf{r}_{1}^{'}$, $\mathbf{r}_{2}^{'}$ is the MLP output features of rays from $\mathbf{Ray}_{i,n}^{'}$ and their neighbour region $\mathcal{R}^{'}$.
Besides, $\phi, \ell_{jsd}$ is the softmax function and Jensen-Shannon Divergence (JSD), respectively.

\textbf{Temporal Invariant Consistency:} Which aims to provide geometric consistency constraints between adjacent frames in RGB sequences.
For each image in the input RGB sequence, we construct a series of image pairs. 
Each pair consists of the current frame and the subsequent frame, which we denote as the matched images.
We extract SIFT features from all images and then identify SIFT correspondences in their other matched image.
Through this process, we determine the number of temporal invariant consistency correspondences and their spatial locations.
\begin{figure}[t]
\centering
{\includegraphics[width=7cm]{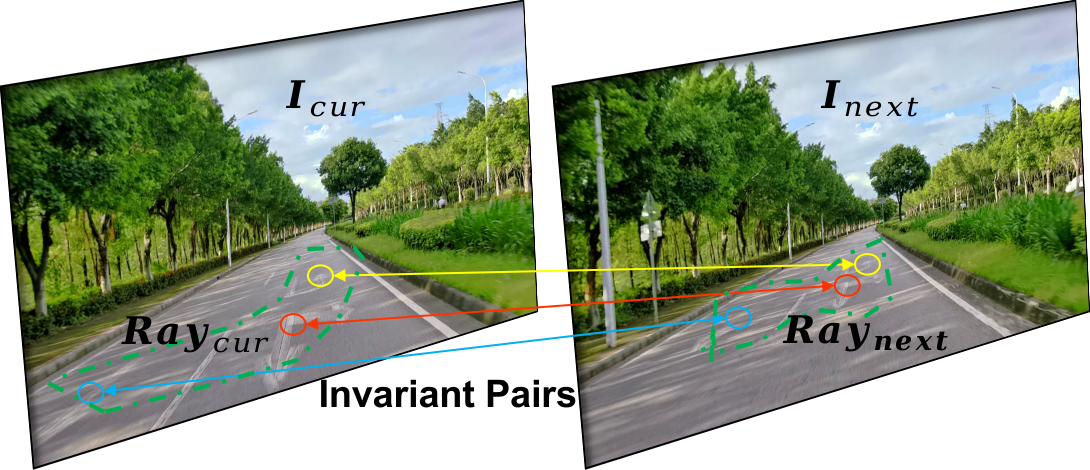}}\hspace{0.1mm}
\caption{
The temporal invariant consistency loss mitigates point cloud artifacts by ensuring the representation consistency of SIFT correspondence rays between two matched images.
In the image above, distinct colored lines denote various SIFT correspondences, while the small circle indicates the pixel position corresponding to the matching rays.}
\label{tic}
\end{figure}
Given a pair of images $\{\mathbf{I}_{cur}, \mathbf{I}_{nex} \in \mathbb{R}^{H \times W \times 3} \}$, where $\mathbf{I}_{cur}$ corresponds to the current frame and $\mathbf{I}_{next}$ represents the subsequent frame, they are a pair of matched image.
Considering these two images, we denote the ray bundles corresponding to their matched SIFT features as $\mathbf{Ray}_{cur}, \mathbf{Ray}_{next}$. 
Each ray bundle comprises N pairs of correspondences.
Inspired by the spatial dynamic consistency loss $\ell_{sdc}$, we aim to ensure similarity between features of matched SIFT points in adjacent temporal frames.
To achieve this, we introduce a new loss term called temporal invariant consistency loss $\ell_{tic}$.
The calculation of $\ell_{tic}$ involves the Jensen-Shannon Divergence function as introduced in the calculation of $\ell_{sdc}$.
Specifically, we compute it as follows: 
\begin{equation}
\begin{aligned}
\ell_{tic} &= \ell_{jsd} (\phi(\mathbf{r}_{cur}^{'}),\phi(\mathbf{r}_{next}^{'}))\\
&= 0.5 * \frac{\sum \phi(\mathbf{r}_{cur}^{'}) * ln(\phi(\mathbf{r}_{next}^{'})) + \sum \phi(\mathbf{r}_{next}^{'}) * ln(\phi(\mathbf{r}_{cur}^{'}))}{(\phi(\mathbf{r}_{cur}^{'}) + \phi(\mathbf{r}_{next}^{'}))/2},
\label{tic}
\end{aligned}
\end{equation}
where $\mathbf{r}_{cur}^{'}$, $\mathbf{r}_{next}^{'}$ represent the MLP output features of rays from $\mathbf{Ray}_{cur}$ and $\mathbf{Ray}_{next}$, respectively.

\subsubsection{Total Losses}\label{sec:total_loss}
The total loss for our proposed NeRF2Points model is given by:
\begin{equation}
\begin{aligned}
\ell_{total} &= \ell_{rec}(.) + \beta * \ell_{sdc}(.) + \gamma * \ell_{tic}(.),\\
\ell_{rec} &= \theta_{1} * \ell_{rgb}(.) + \theta_{2} * \ell_{depth}(.) + \theta_{3} * \ell_{normal}(.),
\end{aligned}
\end{equation}
where $\theta_{1}, \theta_{2}, \theta_{3}, \beta, \gamma$ represent the balanced coefficients, set to 1.0, 0.1, 0.005, 0.01, and 0.01, respectively.
The reconstruction loss $\ell_{\text{rec}}$ captures the error in predicting RGB, depth, and normal information for a batch of ray bundles. 
These three loss terms are detailed in Eq. \ref{cost}.
The spatial dynamic consistency loss $\ell_{\text{sdc}}$ quantifies spatial consistency. 
It is computed as described in Eq. \ref{sdc}.
The temporal invariant consistency loss $\ell_{\text{tic}}$ is based on SIFT correspondences between a pair of matched images, as expressed in Eq. \ref{tic}.

\section{Experimentals}
\subsection{Implementation Details}
Our NeRF2Points framework extends NeRFacto \cite{zhang2021nerfactor} within the NeRFSTUDIO ecosystem \cite{tancik2023nerfstudio}, adhering to a training regimen akin to the implementation of NeRFacto in NeRFSTUDIO:  
We employ a scene space annealing approach in the initial training stages, complemented by an exponential learning rate decay with a warm-up phase. 
The optimization is conducted using the Adam algorithm, reinforced by gradient clipping set to a threshold and norm of 0.1.  
NeRF2Points undergoes a training regimen of 20,000 iterations with a batch size of 2048 on a single NVIDIA A6000 GPU. The the parameters $\lambda_{1}, \lambda_{2}$, and $\lambda_{3}$, crucial for cost matrix computation in Section \ref{sec:GAC}, are calibrated to 0.1, 1.0, and 0.05, respectively.

\subsection{Datasets and Metrics}

Our experimental analysis leverages a proprietary dataset detailed in Sec. \ref{sec:svdc}, consisting of roughly 180,000 high-definition street-view images sourced from autonomous vehicles over a 20-kilometer span.  
To ascertain the precision of camera poses within our driving dataset, we utilize initial poses derived from Wheel-IMU-GNSS Odometry (WIGO), which are subsequently refined through a structure-from-motion workflow.  

For image quality assessment, we compute the average PSNR (Peak Signal-to-Noise Ratio) and SSIM (Structural Similarity Index). 
In evaluating the fidelity of the generated point clouds, we apply the Chamfer Distance (C-D) metric to gauge geometric congruence with ground-truth point clouds obtained from LiDAR sensors, a standard measure in point cloud completion endeavors:
\begin{equation}
\textit{C-D} (\hat{\mathbf{PC}}, \mathbf{PC}) = \frac{1}{| \hat{\mathbf{PC}}|} \sum_{\mathbf{x} \in \hat{PC}} \min_{\mathbf{y} \in \mathbf{PC}} \lVert \mathbf{x}-\mathbf{y} \rVert_2^2 + \frac{1}{|\mathbf{PC|}} \sum_{\mathbf{y} \in \mathbf{PC}} \min_{\mathbf{x} \in \hat{\mathbf{PC}}} \lVert \mathbf{y}-\mathbf{x} \rVert_2^2,
\end{equation}
where $\mathbf{PC}, \hat{\mathbf{PC}}$ are generated and LiDAR sensor provides point clouds, $\mathbf{x}$ and $\mathbf{y}$ represent the three-dimensional coordinates in these two point clouds.

\begin{figure}[htp]
\centering
\subfloat{\includegraphics[width=10cm, height=3cm]{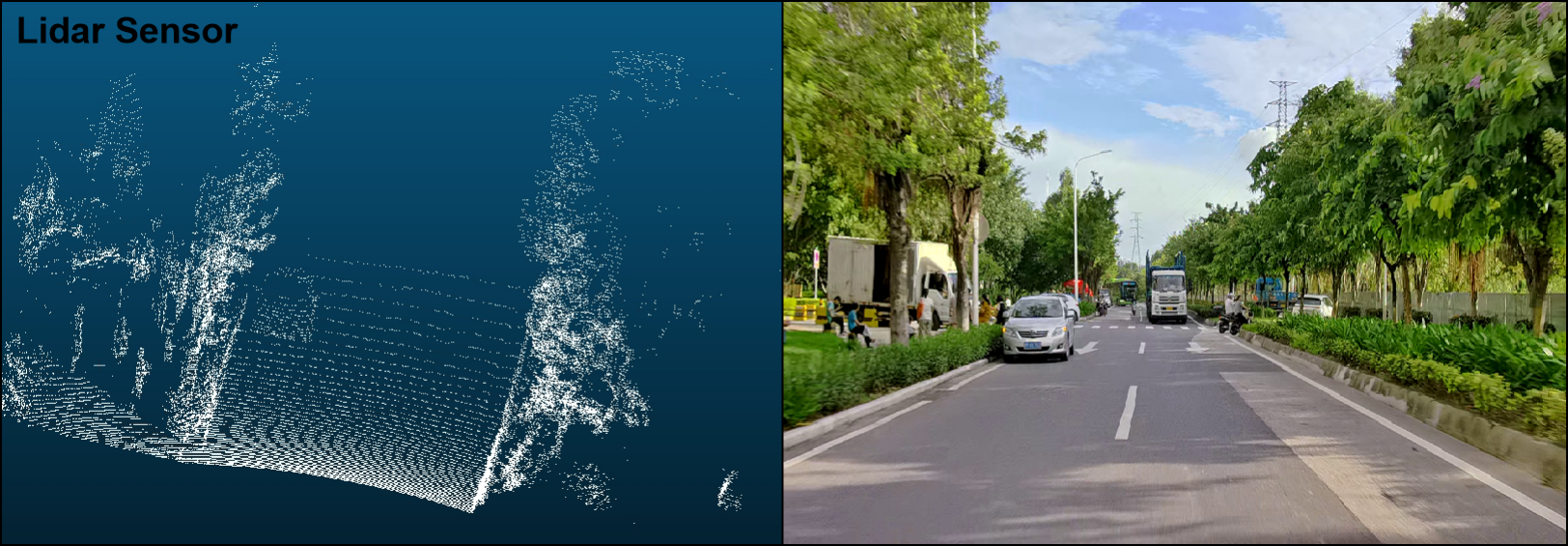}}\\
\subfloat{\includegraphics[width=10cm, height=5cm]{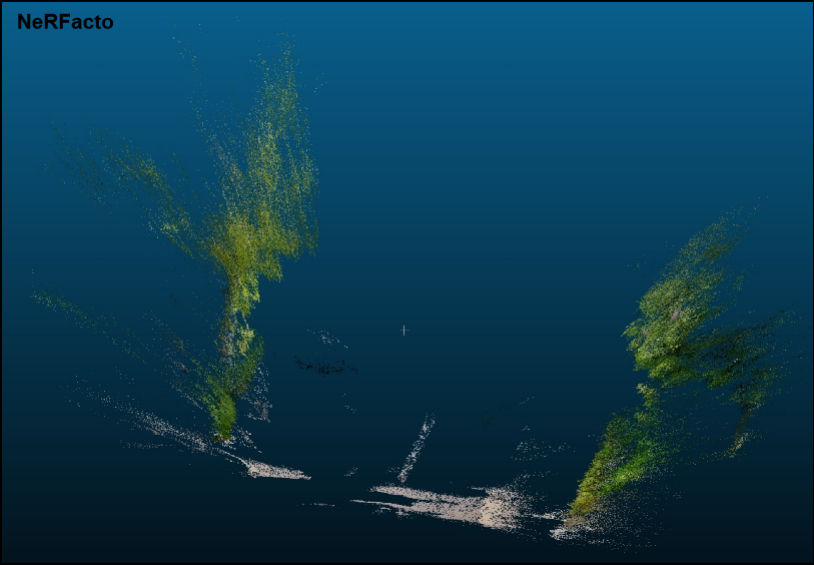}}\\
\subfloat{\includegraphics[width=10cm, height=5cm]{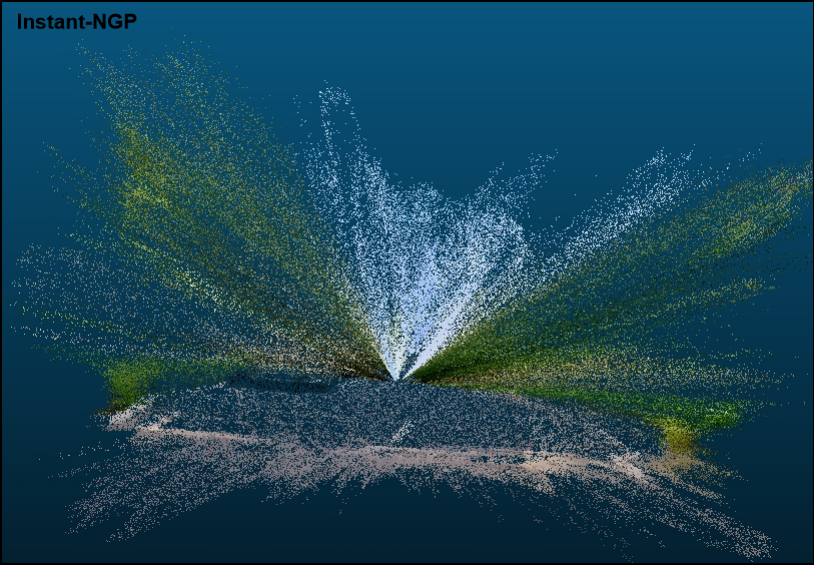}}\\
\subfloat{\includegraphics[width=10cm, height=5cm]{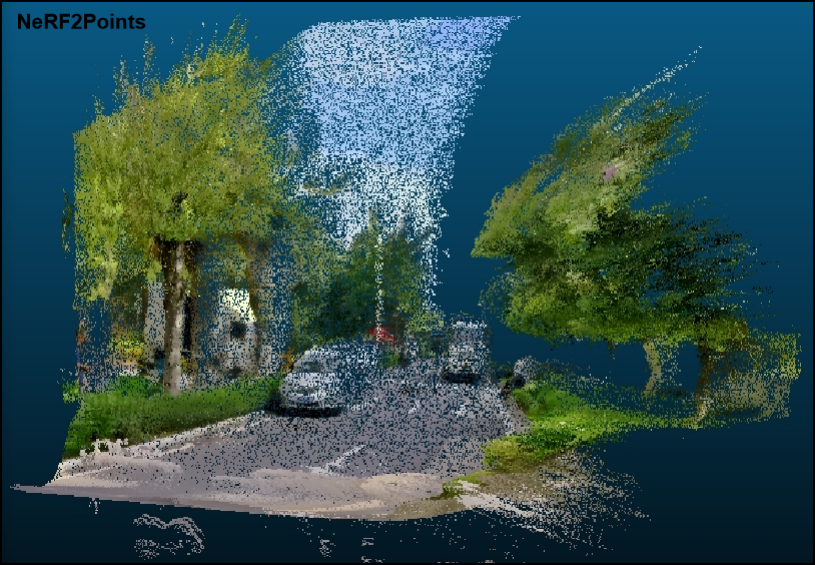}}
\caption{The generated point cloud corresponding to a single-frame image is depicted.
The top row showcases the input RGB image alongside its corresponding point cloud obtained from the Lidar sensor.
Subsequent rows display RGBD point clouds reconstructed using various methods, all based on the same single-frame image.}
\label{vis_compare_sote}
\end{figure}
\subsection{Quantitative Results}
In this section, we conduct a comparative analysis between NeRF2Points and five cutting-edge NeRF-based methods: NeRFacto \cite{zhang2021nerfactor}, Instant-NPG \cite{muller2022instant}, NeuS \cite{wang2021neus}, Depth-NeRFacto \cite{zhang2021nerfactor} and Mip-NeRF \cite{barron2022mip}. 
We present their performance across three evaluation metrics in Table \ref{compare_sota}.
While Mip-NeRF excels in object-centered outdoor reconstruction, it falls short when dealing with street view data.
The second row of Fig. \ref{vis_compare_sote} illustrates that NeRFacto’s performance on street view data is terrible, with a reconstructed RGBD point cloud road lacking distinct features.
Despite Instant-NPG achieving the highest image rendering metrics for street view data among these five methods, street view details remain largely invisible in the RGBD point cloud it produces as shown in the third row of Fig. \ref{vis_compare_sote}.
In contrast, NeRF2Points, on the other hand, can produce clear RGBD point clouds while maintaining impressive image rendering performance.
It’s important to highlight that NeRF2Points retains the ability to reconstruct the visual appearance of a moving vehicle while reconstructing the point cloud from a single-frame image (the fourth of Fig. \ref{vis_compare_sote}).
\begin{table*}[htp]
\centering
\caption{Comparison of NeRF2Points and five other NeRF-base SOTA methods.}
\begin{tabular}{c|c|c|c}
\hline
    \multirow{2}{*}{Methods} & \multicolumn{3}{c}{Results}\\\cline{2-4}
         & $~~PSNR~~~\uparrow$ &  $~~SSIM~~~\uparrow$ & $~~C-D~~~\downarrow$\\\hline
    NeRFacto \cite{zhang2021nerfactor} & 19.788 & 0.560 &  5.617 \\
    Depth-NeRFacto \cite{zhang2021nerfactor} & 19.653 &  0.524 & 5.881  \\
    Instant-NPG \cite{muller2022instant} &  24.169 & 0.812 & 5.035 \\ 
    Mip-NeRF \cite{barron2022mip} &  17.723 & 0.420 & 6.690  \\
    NeuS \cite{wang2021neus} & 21.071 & 0.567 &  5.327 \\\hline
    NeRF2Points & 29.139 & 0.831 & 1.876 \\
    \hline\hline
    
\end{tabular}
\label{compare_sota}
\end{table*}

\begin{table*}[htp]
\centering
\caption{Ablations on different settings.
The second row illustrates the point cloud reconstruction achieved without certain components from NeRF2Points.
In contrast, the third row showcases the point cloud reconstruction when all components of NeRF2Points are utilized.}
\begin{tabular}{c|c|c|c}
\hline
    \multirow{2}{*}{} & \multicolumn{3}{c}{Results}\\\cline{2-4}
         & $~~PSNR~~~\uparrow$ &  $~~SSIM~~~\uparrow$ & $~~C-D~~~\downarrow$\\\hline
    w/o LiPM   & 20.861 &  0.681 & 4.600  \\
    w/o $\ell_{sdc}$ &  25.15 & 0.784 & 2.035 \\ 
    w/o $\ell_{tic}$ &  28.549 & 0.820 & 1.901  \\\hline
    Full Settings & 29.139 & 0.831 & 1.876 \\
    \hline\hline
    
\end{tabular}
\label{ablation_studies}
\end{table*}
\subsection{Ablation Studies}
To further demonstrate the effectiveness of our method, we conduct a series of ablation studies with or without certain components in Table. \ref{ablation_studies}.
The table indicates that LiPM significantly affects NeRF2Points’ performance by directly influencing the radiation field’s ability to model the road surface.
Next, the loss $\ell_{sdc}$ is followed by spatial plane optimization, which mitigates point cloud artifacts.
While $\ell_{tic}$ does constrain the radiation field learning process, its impact on performance improvement is minimal.

\subsection{Conclusion}
This paper introduces a novel task: generating RGBD point clouds from RGB sequences using NeRF technology, referred to as NeRF2Points.
Our experiments reveal that NeRF2Points can create RGBD point clouds that remain clear structures.
Interestingly, NeRF2Points can reconstruct the visual appearance of a moving vehicle from a single-frame image.
In the future, we will investigate 4D point cloud scene reconstruction using NeRF2Points.

\section*{Appendix}
\appendix
\begin{figure}[htp]
\centering
\subfloat[inputs]{\includegraphics[width=2.5cm]{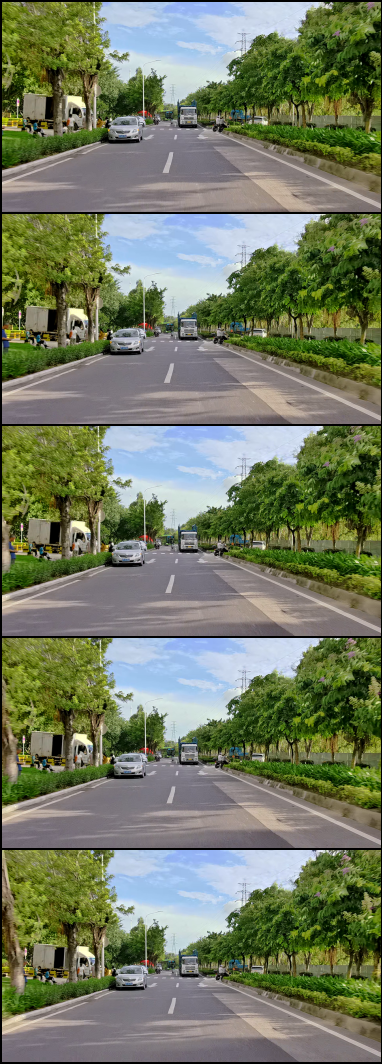}}
\subfloat[street-views]{\includegraphics[width=2.5cm]{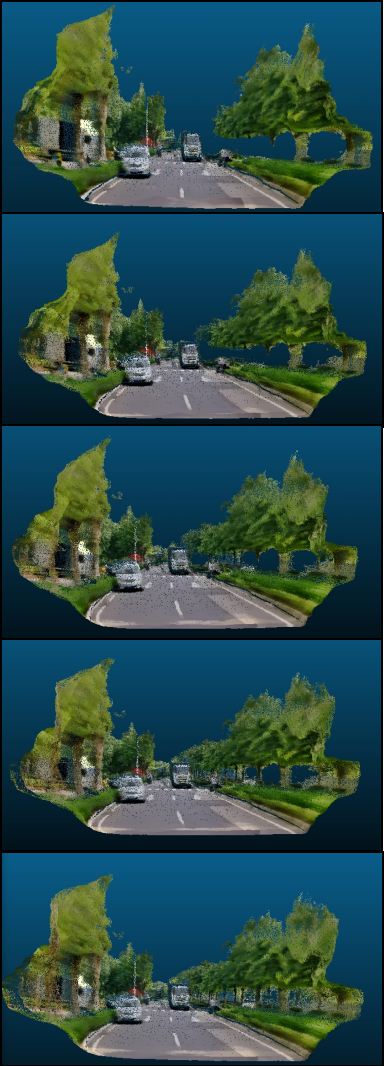}}
\subfloat[road]{\includegraphics[width=2.5cm]{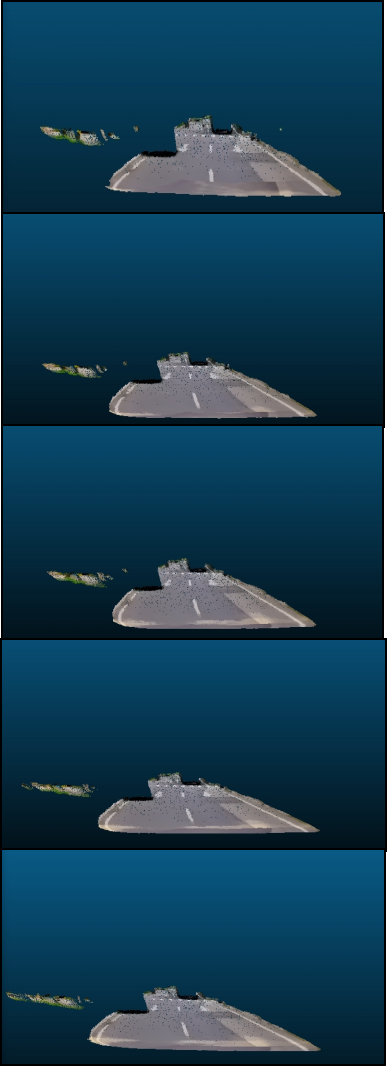}}
\subfloat[scenes]{\includegraphics[width=2.5cm]{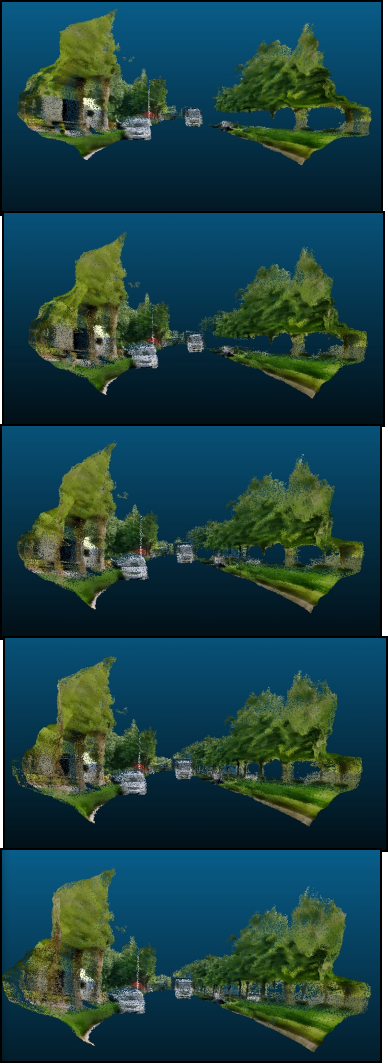}}
\subfloat[moving cars]{\includegraphics[width=2.5cm]{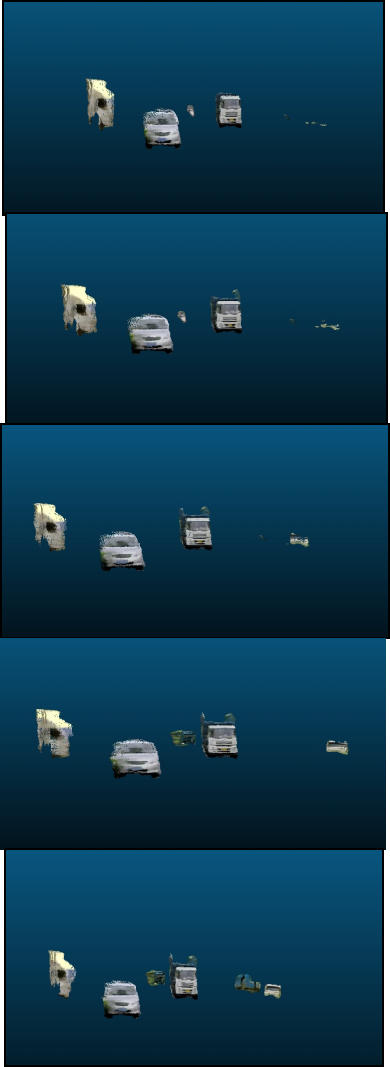}}
\caption{More visualization of generated point clouds of NeRF2Points.}
\label{vis_compare_sote}
\end{figure}
\subsection{More visualization}
In Fig. 1, we show the corresponding point cloud generated by NeRF2Points from an RGB sequence of five consecutive frames. 
In this process, we use a trained image semantic segmentor to make NeRF2Points possible to generate point clouds of various regions of interest separately.

\subsection{Point Cloud Generation}
After learning the implicit NeRF representation of the street-views, we produce RGBD point clouds from trained radiance fields.
Denoting the rays of each pixel in input images as ${Rays}$, their corresponding origin and direction were $\mathbf{origin}, \mathbf{direction}$.
Trained radiation fields' depth prediction on $Rays$ is $\mathbf{depth}$, then we can obtained a point cloud by $\mathbf{PC} = \mathbf{origin} + \mathbf{direction} * \mathbf{depth}$.
%
%
\bibliographystyle{splncs04}
\bibliography{main}

\end{document}